\documentclass[conference]{IEEEtran}
\IEEEoverridecommandlockouts

\usepackage{cite}
\usepackage[pdftex]{graphicx}
\usepackage{amssymb}
\usepackage{amsmath}
\usepackage{algorithm}
\usepackage{algpseudocode}
\usepackage{tabularx}
\usepackage{booktabs}
\usepackage{multirow}
\usepackage{array}
\newcolumntype{P}[1]{>{\raggedright\arraybackslash}p{#1}}
\usepackage{float}
\usepackage[hyphens]{url}
\usepackage{hyperref}
\begin{document}

\title{Federated learning with hierarchical clustering of local updates to improve training on non-IID data\\
\thanks{This work is supported by the SEND project (grant ref. 32R16P00706) funded by ERDF and BEIS.}
}

\author{\IEEEauthorblockN{Christopher Briggs}
\IEEEauthorblockA{\textit{School of Computing \& Mathematics} \\
\textit{Keele University}\\
Keele, UK \\
c.briggs@keele.ac.uk}
\and
\IEEEauthorblockN{Zhong Fan}
\IEEEauthorblockA{\textit{School of Computing \& Mathematics} \\
\textit{Keele University}\\
Keele, UK \\
z.fan@keele.ac.uk}
\and
\IEEEauthorblockN{Peter Andras}
\IEEEauthorblockA{\textit{School of Computing \& Mathematics} \\
\textit{Keele University}\\
Keele, UK \\
p.andras@keele.ac.uk}
}

\maketitle

\begin{abstract}
  Federated learning (FL) is a well established method for performing machine learning tasks over massively distributed data. However in settings where data is distributed in a non-iid (not independent and identically distributed) fashion - as is typical in real world situations - the joint model produced by FL suffers in terms of test set accuracy and/or communication costs compared to training on iid data. We show that learning a single joint model is often not optimal in the presence of certain types of non-iid data. In this work we present a modification to FL by introducing a hierarchical clustering step (FL+HC) to separate clusters of clients by the similarity of their local updates to the global joint model. Once separated, the clusters are trained independently and in parallel on specialised models. We present a robust empirical analysis of the hyperparameters for FL+HC for several iid and non-iid settings. We show how FL+HC allows model training to converge in fewer communication rounds (significantly so under some non-iid settings) compared to FL without clustering. Additionally, FL+HC allows for a greater percentage of clients to reach a target accuracy compared to standard FL. Finally we make suggestions for good default hyperparameters to promote superior performing specialised models without modifying the the underlying federated learning communication protocol.
\end{abstract}


\IEEEpeerreviewmaketitle

\section{Introduction \& Background}

The machine learning setting known as federated learning (FL) \cite{2019arXiv191204977K, YangQiang:2019bx, 2019arXiv190807873L} allows for local learners to participate in training a joint statistical model $f(w)$ on massively distributed data. This is useful in situations where data cannot be gathered into a single location for legal or privacy reasons.

The most common setting for FL employs a central entity for aggregating local learner models from many clients. These clients are typically mobile and Internet-of-Things devices. During training, a fraction of the clients are sent the current state of the global model $w_t$. Using only the data available to the client, a training procedure (defined by the central entity) is run locally to work towards minimising some local objective $F_k(w_t)$. For example, training a neural network model for image classification, the stochastic gradient descent (SGD) algorithm could be run on the client to iteratively minimise the classification error over all training images available to the client. 

\begin{equation}
    w_{t+1}^{k} \leftarrow w_t - \eta \nabla F_k(w_t).
\end{equation}

The resulting model $w_{t+1}^k$ from each client $k \in K$ is sent back to the central entity where all local models are aggregated in some fashion. The commonly used FederatedAveraging algorithm \cite{mcmahan2017communication} performs a data-weighted average over the parameters of all models to form a new global model $w_{t+1}$ ready for the next round of training ($n_k/n$ is the number of samples available to client k versus all samples).

\begin{equation}
	w_{t+1} \leftarrow \sum_{k=1}^{K} \frac{n_k}{n} w_{t+1}^k.
\end{equation}

Clients can perform one or multiple steps of gradient descent before sending weight updates as orchestrated by the federated algorithm.

FL has been shown to work well in situations where the data is distributed in an independent and identically distributed (iid) fashion. However, in a typical scenario where data is massively distributed among clients, that data is likely to be unbalanced and non-iid. In this work we explore different non-iid data distributions and apply a hierarchical clustering algorithm to determine client similarity. In this way we can train multiple disjoint models that are targeted to clusters of similar clients with the intention to improve the performance of the training objective for all clients whilst reducing communication in the FL protocol.

\subsection{Contributions}
Our contributions through this work include:
\begin{itemize}
    \item A method for training specialised models for subsets of clients that can increase test set accuracy whilst reducing the number of communication rounds required to reach convergence. This is achieved by clustering the clients by similarity based on their updates to the global joint model after a set number of communication rounds.
    \item A full characterisation of how hierarchical clustering affects test set accuracy when applied during FL training in iid and various non-iid settings.
    \item An empirical analysis of the effect that varying hyperparameters for FL and the hierarchical clustering algorithm has on forming good specialised models for subsets of clients.
    \item Recommendations for good default hyperparameters to use for training with our method when the non-iid nature of the data is unknown (as is likely to be the case when using FL).
\end{itemize}

\subsection{Client statistical heterogeneity}
One of the fundamental challenges for training a single joint model under FL is the presence of non-iid data. There should be no assumption that clients have access to data drawn independently from the same underlying distribution - $\mathcal{P}_i \neq \mathcal{P}_j$ for all pairs of clients $i$ and $j$. Unlike under the iid assumption the local model $F_k$, in expectation, trained on its own data $D_k$ will be an unreliable approximation of the joint model $f$:

\begin{equation}
	\mathbb{E}_{D_k}[F_k(w)] \neq f(w)
\end{equation}

There are a variety of ways in which the data among clients can be distributed in a non-iid fashion \cite{2019arXiv191204977K}: 

\begin{itemize}
    \item \textbf{Feature distribution skew*}: The $\mathcal{P}_i(x)$ marginal distributions vary between clients. This is the case where the input features are not evenly distributed between clients.
    \item \textbf{Label distribution skew*}: The $\mathcal{P}_i(y)$ marginal distributions vary between clients. This is the case where the data labels are not evenly distributed between clients. For example some clients only have access to data from a subset of all possible labels for a given task.
    \item \textbf{Concept shift (same features, different label)*}: The $\mathcal{P}_i(y|x)$ conditional distributions vary between clients. Here different labels are assigned for the same features across clients. For example, client $i$ labels all cat images as 'cat', but client $j$ labels all cat images as 'dog'.
    \item \textbf{Concept shift (same label, different features)}: The $\mathcal{P}_i(x|y)$ conditional distributions vary between clients. Here different features across clients are labelled with the same label. For example, client $i$ labels all cat images as 'cat', but client $j$ labels all dog images as 'cat'.
\end{itemize}

Additionally, clients might have access to varying numbers of training examples. In practice, some degree of all of these non-iid settings may be represented in a given massively distributed dataset. In this work we explore the effect of different non-iid distributions on the ability for hierarchical clustering to determine client similarity from their client updates, namely the starred (\textbf{*}) non-iid settings above.

Where data is distributed in a highly non-iid fashion, a single joint model may not be able to meet the objectives of all clients simultaneously \cite{2019arXiv191001991S}. Instead, multiple models targeted towards groups of clients with similar data distributions might be preferred. However in FL, the raw data cannot be inspected by the central entity. As such, only the local model updates can be used to judge the direction in which each client wishes to update the global model. This provides the only proxy for determining client similarity whilst preserving some level of privacy over the local data.

\subsection{Hierarchical clustering}
Training multiple models for subsets of related clients can be achieved by clustering the model updates received from the clients. Many unsupervised clustering algorithms require an estimation of the number of these clusters a priori. Since we cannot know how many unique data generating distributions the client's datasets are drawn from, a clustering algorithm that can determine the number of clusters independently must be employed. However, some clustering methods that determine the number of clusters automatically fail to assign outlying samples to a cluster and simply label them as noise (for example DBSCAN \cite{Ester:1996tm}). Hierarchical clustering \cite{Hastie:2009fg} is a natural choice for the purpose of clustering where the number of clusters is unknown and where all examples are assigned to the most relevant cluster. Another benefit of using hierarchical clustering is its ability to scale to large numbers of samples and clusters as well as being reasonably interpretable.

In this work we opt to use an agglomerative hierarchical clustering method which begins with all samples belonging to their own singleton cluster. Each sample is simply a vectorised local model update (the parameters of the local model). At each step of the clustering, the pairwise distance between all clusters is calculated to judge their similarity. The two clusters that are most similar are merged. This continues for a total of $N-1$ steps until a single cluster remains, containing all the samples. Thus, a hierarchical structure of similarity between clients is constructed. This structure possesses the property of monotonicity, such that the dissimilarity of two clusters involved in merging increases with successive steps. A distance threshold can act as a hyperparameter to determine when to stop merging clusters. Intuitively, this acts to limit how dissimilar two clusters can be before stopping the merging process.

Two further hyperparameters of the hierarchical clustering algorithm are of importance to our study. The first is the distance metric uses to compute the similarity between clusters. In this work we opt to test L1 (Manhattan), L2 (Euclidean) and cosine distance metrics. Euclidean distance is a common metric used to judge similarity between vectors, however the Manhattan distance works well for sparse vectors and is less affected by outliers. Finally, the cosine distance metric is invariant to scaling effects and therefore only indicates how closely two vectors point in the same direction.

The second important hyperparameter is the linkage mechanism for determining how similar two clusters are. Single linkage determines distance based on the most similar pair of samples across two clusters. Complete linkage determines distance based on the most dissimilar pairs of samples across two clusters. Average linkage averages the samples within each cluster and compares distances based on these averages. Finally Ward's linkage (which can only be combined with the Euclidean distance metric) seeks to minimise the intra-cluster variance upon merging two clusters.

Our experiments test the ability of combinations of these hyperparameters to produce clusters of client updates that, when averaged, exhibit a higher test set accuracy compared to training using a single joint model (as in the normal FL setting).

\subsection{Related work}
The FL setting \cite{mcmahan2017communication} was envisioned with non-iid data in mind unlike previous works on distributed training \cite{McDonald:2014en, Povey:2014us} which focused on iid data in a data center environment. The commonly used FederatedAveraging algorithm \cite{mcmahan2017communication} makes no special adjustments when encountering non-iid data and therefore suffers a penalty in performance and/or communication costs in such circumstances \cite{Hsieh:2019vg}. Several works have sought to deal with non-iid data in a FL setting. Zhao et al. \cite{2018arXiv180600582Z} propose sharing a small subset of public data with heterogenous clients to reduce the weight divergence between trained local models, thus increasing rubustness and stability during training. However this relies on a substantial amount of public data being available for a given task. Li et al. \cite{2018arXiv181206127L}, solve the same problem by adding a regularisation term to the local optimisation, limiting the weight divergence of the local and global models. Both of these approaches train a single joint model and work well to increase overall model accuracy. However, under some non-iid data distributions (particularly where clients have competing objectives), a single joint model cannot perform optimally for all clients simultaneously \cite{2019arXiv191001991S}. For the purpose of improving performance when faced with client datasets across different timezones, Eichner et al. \cite{2019arXiv190410120E} propose a semi-cyclic method to train pluralistic models that perform model averaging over blocks of clients (e.g. clients in particular timezones). In this work clients are clustered where they are located in a shared timezone. Our approach does not require any extra knowledge about the clients (timezones or otherwise) as only the local updates are observed and used to form the clusters automatically.

By considering each local learning objective as a task, multi-task learning \cite{2017arXiv170708114Z} has been employed in the distributed learning setting in multiple works \cite{Liu:2017dz, Wang:vg, 2017arXiv170510467S}. Another multi-task learning setup considers the learning objectives of a subset of related clients as a single task. Ghosh et al. \cite{2019arXiv190606629G} use K-means to cluster client updates for the purpose of identifying and isolating byzantine clusters of clients prior to the averaging step. Our work most closely resembles that of Sattler et al. \cite{2019arXiv191001991S} who develop a clustered federated learning (CFL) procedure. This uses an optimal bipartitioning algorithm to separate clients based on cosine similarity for the purpose of producing specialised models for each cluster of clients. In contrast, our work covers a wider range of non-iid settings, similarity measures and allows for training on only a fraction of clients per round of communication. We also employ a single clustering step compared to possibly many in CFL, reducing the load of the clustering procedure on the aggregation server.
\section{Federated learning with hierarchical clustering}

Under the FL setting, the assumption is that the objectives for all clients approximate the global objective. However, in the presence of non-iid data, this is not the case. As such, we propose a federated learning with hierarchical clustering (FL+HC) setting. During the FL procedure, a clustering step at communication round $n$ is introduced. At the clustering step, we perform a communication round involving all the clients on the global joint model trained up to round $n$. The updated local model from all clients is used to judge the similarity between clients and the hierarchical clustering algorithm is employed to iteratively merge the most similar clusters of clients up to a given distance threshold $T$. Once merging is halted, the determined clusters of clients $C$ are trained independently but simultaneously, initialised with the joint model at its current state. This algorithm is detailed in algorithm \ref{alg:FLHC}. If a good clustering has been achieved, the average test set accuracy over all clients should outperform that of training a single joint model under FL. Under FL+HC we are training a specialised model $f_c(w)$ for every cluster of similar clients $c$, each of which have different objectives such that:

\begin{equation}
	\forall{c \in C}, \mathbb{E}_{D_k}[F_k(w)] = f_c(w) \text{ where } k \in c
\end{equation}

The operation required to perform the hierarchical clustering runs in $\mathcal{O}(n^3)$ time. As the operation occurs on the server which is assumed to be large enough to aggregate the updates from many clients, and only occurs once during training, the impact on the overall training operation is not significant.

\begin{algorithm}[h]
	\caption{\textsc{Federated learning with hierarchical clustering (FL+HC)}. $n$ is the number of rounds of FL prior to clustering, $\alpha$ is the fraction of clients selected to participate in each round of FL and $P$ is the set of hyperparameters for the hierarchical clustering algorithm. The $K$ clients are indexed by $k$ and $C$ discovered clusters are indexed by $c$. $K_c$ is the set of clients in cluster $c$. On the client, $B$ is the local mini-batch size, $\mathcal{P}_k$ is the dataset available to client $k$, $E$ is the number of local epochs, and $\eta$ is the learning rate}
	\label{alg:FLHC}
    \begin{algorithmic}[1]
    \Procedure{FL+HC}{}\Comment{On server}
        \State Initialise $w_0$
        \For {each round $t \in [1, n]$}
            \State $w_{t+1} \gets$ \textsc{FederatedLearning}($w_t, K$)
        \EndFor
        \State $w \gets w_{t+1}$
        \For {each client $k \in K$}\Comment{In parallel}
            \State $\Delta w^k \gets$ \textsc{ClientUpdate}($k, w$)
        \EndFor
        \State $C \gets$ HierarchicalClusteringAlgorithm($\Delta w, P$)
        \For {$c \in C$}\Comment{In parallel}
            \State $w_{c,0} \gets w$
            \For {each round $t = 1, 2, \dots$}
                \State $w_{c,t+1} \gets$ \textsc{FederatedLearning}($w_{c,t}, K_c$)
            \EndFor
        \EndFor
	\EndProcedure
	\Statex
	\Procedure{FederatedLearning}{$w_t, K$}\Comment{On server}
        \State $m \gets \max{(\alpha \cdot K, 1)}$
        \State $S_t \gets$ (random set of $m$ clients)
        \For {each client $k \in S_t$}\Comment{In parallel}
            \State $w_{t+1}^k \gets$ \textsc{ClientUpdate}($k, w_t$)
        \EndFor
        \State $w_{t+1} \gets \sum_{k=1}^K \frac{n_k}{n} w_{t+1}^k$
	\EndProcedure
	\Statex
	\Procedure{ClientUpdate}{$k, w$}\Comment{On client $k$}
	\State $\mathcal{B}\gets$ (Split $\mathcal{P}_k$ into batches of size $B$)
	\For {each local epoch $i$ from 1 to $E$}
		\For {batch $b \in \mathcal{B}$}
			\State $w \gets w - \eta \nabla \mathcal{L}(w;b)$
		\EndFor
	\EndFor
	\State return $w$ to server
	\EndProcedure
	\end{algorithmic}
\end{algorithm}

\subsection{Experiment setup}

Our work focuses on the computer vision image classification problem of identifying handwritten digits from pixel data. The well-known MNIST dataset \cite{Lecun:1998hy} was chosen for this purpose as it has been studied by various FL researchers \cite{mcmahan2017communication,2019arXiv191001991S,2018arXiv181206127L} and provides a simple enough task to test various clustered settings and data partitions. The MNIST dataset contains 60,000 training examples and 10,000 test examples.

The data partitions were designed specifically to target various ways in which data distributions might differ between clients. The iid setting shuffles all the data and divides it among 100 clients evenly (600 training examples each). In this scenario, $\mathcal{P}_i = \mathcal{P}_j$ for all pairs of clients $i$ and $j$. The first non-iid setting, referred to as pathological non-iid (as described in \cite{mcmahan2017communication}) partitions the data such that clients receive digits corresponding to only 2 labels. For example the first client might receive 300 examples labelled as 2 and 300 examples labelled as 7. Subsequent clients might receive different labels. Again all clients have 600 examples to perform local learning on. In this setting, we test how the FL model performs with label distribution skew, as $\mathcal{P}(y)$ varies across clients. 

The next non-iid setting, referred to as label-swapped non-iid (described in \cite{2019arXiv191001991S}) first shuffles the data and then partitions the data into four groups. For each group two digit labels are swapped. For example one group might swap all digits labelled as 3 to 9 and vice versa. Each group is then evenly distributed to 25 clients, resulting in 100 clients each with 600 training examples. This way the clients naturally form 4 clusters and allow us to test FL's ability to train models in the presence of concept shift ($\mathcal{P}(y|x)$ varies between clients). 

The final non-iid setting represents a slightly more challenging task, but a more realistic setting. The FEMNIST dataset \cite{Caldas:2018vk} is used to classify not only 10 handwritten digits but also 26 uppercase and 26 lowercase letters and is pre-partitioned according to the person who wrote the characters. From the original 3500 users in the dataset, 367 users were randomly selected to form the dataset for each of 367 clients. Each client has access to an uneven number of samples (between 12 and 386 samples). This scenario tests FL in the presence of feature distribution skew, where $\mathcal{P}(x)$ varies over the clients as different users write characters in subtly different ways. Additionally, due to the uneven sample sizes, $\mathcal{P}(y)$ will also vary to some degree between clients. The conditional distributions are expected to remain the same in FEMNIST.

The test dataset that each client has access to is always drawn from the same distribution as the training data in all partitioning schemes described above.

To perform machine learning on this dataset, a simple convolutional neural network was designed taking single channel, 28 x 28 pixel images and passing these through two 5x5 convolutional layers (32 and 64 channels respectively) with Relu activations. Each convolutional layer is followed by a 2x2 max pooling layer. Finally the network passes data through a fully connected dense layer with 512 units and Relu activations and provides output via a softmax classification over the 10 possible digit labels (62 in the case of FEMNIST).

FederatedAveraging \cite{mcmahan2017communication} is used to train a global model over 50 training rounds for each data partitioning scheme described above. This is repeated using varying client fractions (0.1, 0.2, 0.5 and 1.0) for each round of training. In these experiments (and all subsequent ones), we use mini-batch stochastic gradient descent on the client with a batch size of 10 over 3 epochs per communication round and set the learning rate to 0.1. The results from these experiments form a baseline against which we compare FL+HC.

For our experiments with applying hierarchical clustering to FL in order to provide better, personalised models targeted to clients with similar updates, we first train a global model for $n$ communication rounds. This global model is then trained for a further 3 epochs on all clients to produce $\Delta w$ (the difference between the global model and local model parameters). The model parameters are reshaped to form a vector and used as feature inputs to the agglomerative hierarchical clustering algorithm. The clustering algorithm returns a number of clusters each containing subsets of clients that are most similar to one another. FL then proceeds for each cluster independently for a total of 50 communication rounds.
\section{Results \& Discussion}

In all our experiments we report the average test set accuracy over all clients in the round directly after the clustering step and at round 50 (where training accuracy begins to plateau using FederatedAveraging with no clustering). At round 50 the model is expected to be close to a stationary solution to the global objective under all settings. We also report the percentage of clients who reach a target of 99\% test set accuracy (80\% for the FEMNIST non-iid setting) in the round after the clustering step and at round 50. Finally, we report the increase/decrease in all these statistics compared to ordinary federated learning with no clustering step. These statistics provide evidence that clustering clients by similarity has a beneficial effect on both the average test set accuracy over all clients and the percentage of clients that perform well at the given task, in most circumstances.

\subsection{Effect of varying client fraction and number of rounds prior to clustering}
In the following experiments we use Ward's linkage mechanism, Euclidean distance metric and fix the distance threshold to 3.0 (10.0 for FEMNIST) to determine when to halt merging of clusters. We varied the fraction of clients participating in each round of training and the number of rounds prior to clustering to see the effect that more rounds of training has on the ability to find good clusters of similar clients. Specifically, the client fraction was tested for 0.1x, 0.2x, 0.5x and 1.0x clients and the number of rounds prior to clustering was tested for 1, 3, 5 and 10 rounds.

In the iid setting, hierarchical clustering was unable to separate clients into clusters for most experiments, instead falling back on the single joint model. This is to be expected as clients with iid partitioned data should return similar updates. However, where the number of rounds prior to clustering was set to just 1 (regardless of client fraction), clusters were identified resulting in a very slightly reduced average test accuracy compared to using the single joint model. This indicates that the stochastic nature of the optimisation on the clients hinders the ability to determine client update similarity until more training iterations have been performed.

\begin{figure}[]
    \centering
    \includegraphics[width=\linewidth]{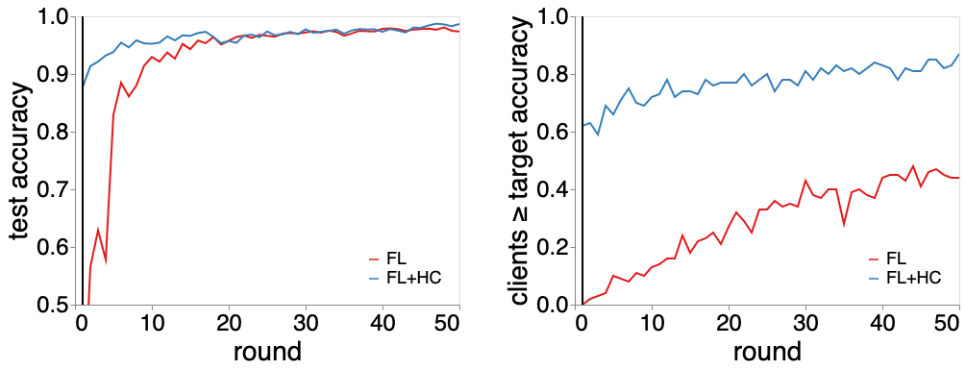}
    \caption{Pathological non-iid setting, using client fraction = 0.2 and number of rounds prior to clustering = 1. The black vertical lines show the round where the clustering step occurs in FL+HC.}
    \label{fig:cf/n r1 pathological non-iid}
\end{figure}

\begin{figure}[]
    \centering
    \includegraphics[width=\linewidth]{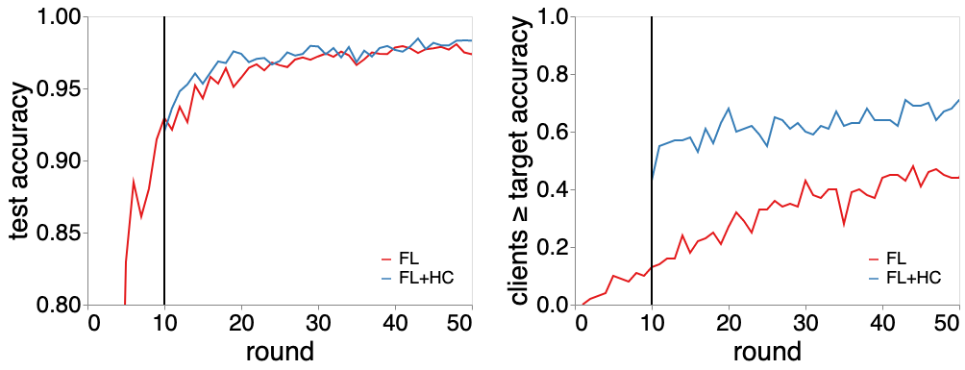}
    \caption{Pathological non-iid setting, using client fraction = 0.2 and number of rounds prior to clustering = 10.}
    \label{fig:cf/n r10 pathological non-iid}
\end{figure}

Under the pathological non-iid setting, there are no strictly pre-defined clusters of similar clients due to the way the data is partitioned. Despite this, we would expect that groups of clients that have access to the same digit labels might produce similar client updates and therefore, some clustering of these updates to produce specialised models may benefit these clients. \autoref{table:cf/n pathological non-iid} shows the results of varying client fraction and the number of rounds prior to clustering for the pathological non-iid setting. In all these results FL+HC outperforms ordinary federated learning. The greatest improvement compared to FL appears to occur where the number of rounds prior to the clustering step is set to 1, resulting in a jump in test set accuracy in the 2nd round of between 1.3x and 1.9x. The final test set accuracy (at round 50) remains the same under FL and FL+HC, but FL+HC asymptotes much more quickly. Similarly, the percentage of clients reaching the target test set accuracy (99\%) exhibits this same jump directly after the clustering step. The percentage of clients reaching the target accuracy is also greatly improved at round 50 compared to FL (1.2x to 2.0x). \autoref{fig:cf/n r1 pathological non-iid} and \autoref{fig:cf/n r10 pathological non-iid} show how test accuracy and the number of client reaching the target accuracy evolves over communication rounds compared to FL. In general, we see a greater improvement in the metrics we are measuring for client fractions $<$1.0.

\begin{table}[]
\caption{FL+HC pathological non-iid setting - cf is the fraction of clients participating in each round of FL, n is the number of rounds of FL prior to clustering. We report the mean test accuracy percentage and the percentage of clients reaching the target accuracy of 99\% directly after the clustering step and at round 50. We also report, in brackets, the increase/decrease in each statistic compared to FL). Values in \textbf{bold} signify the best performance compared to FL in each column.}
\label{table:cf/n pathological non-iid}
\begin{tabularx}{\linewidth}{ccXXXX}
\toprule
                        &    & \multicolumn{2}{c}{test acc}                    & \multicolumn{2}{c}{\% clients}                   \\
cf                   & n  & post cluster           & 50                     & post cluster            & 50                     \\
\midrule
\multirow{4}{*}{0.1} & 1  & \textbf{91.1 (1.9x)} & 98.5 (1.0x)          & 62.0 (20.7x)          & 83.0 (1.8x)          \\
                        & 3  & 80.4 (1.7x)          & 97.1 (1.0x)          & 46.0 (46.0x)          & 67.0 (1.5x)          \\
                        & 5  & 89.3 (1.1x)          & 97.9 (1.0x)          & 45.0 (7.5x)           & 74.0 (1.6x)          \\
                        & 10 & 92.0 (1.1x)          & 98.2 (1.0x)          & 40.0 (2.9x)           & 77.0 (1.7x)          \\
\midrule
\multirow{4}{*}{0.2} & 1  & 91.4 (1.6x)          & \textbf{98.7 (1.0x)} & 63.0 (31.5x)          & \textbf{87.0 (2.0x)} \\
                        & 3  & 92.5 (1.6x)          & 98.3 (1.0x)          & 58.0 (14.5x)          & 73.0 (1.7x)          \\
                        & 5  & 90.2 (1.0x)          & 98.4 (1.0x)          & 49.0 (5.4x)           & 74.0 (1.7x)          \\
                        & 10 & 93.6 (1.0x)          & 98.3 (1.0x)          & 55.0 (3.9x)           & 71.0 (1.6x)          \\
\midrule
\multirow{4}{*}{0.5} & 1  & 91.9 (1.5x)          & 98.7 (1.0x)          & \textbf{68.0 (68.0x)} & 89.0 (1.7x)          \\
                        & 3  & 94.6 (1.1x)          & 98.8 (1.0x)          & 67.0 (16.8x)          & 83.0 (1.6x)          \\
                        & 5  & 93.6 (1.0x)          & 98.3 (1.0x)          & 61.0 (7.6x)           & 78.0 (1.5x)          \\
                        & 10 & 93.8 (1.0x)          & 98.0 (1.0x)          & 39.0 (2.0x)           & 63.0 (1.2x)          \\
\midrule
\multirow{4}{*}{1.0}   & 1  & 94.1 (1.3x)          & 98.6 (1.0x)          & 64.0 (32.0x)          & 77.0 (1.6x)          \\
                        & 3  & 94.3 (1.1x)          & 98.5 (1.0x)          & 64.0 (8.0x)           & 81.0 (1.7x)          \\
                        & 5  & 93.6 (1.0x)          & 99.1 (1.0x)          & 61.0 (5.1x)           & 82.0 (1.7x)          \\
                        & 10 & 95.6 (1.0x)          & 98.8 (1.0x)          & 48.0 (2.8x)           & 70.0 (1.4x)\\
\bottomrule
\end{tabularx}
\end{table}

\begin{table}[]
\caption{FL+HC label-swapped non-iid setting - Empty values are due to 0 clients reaching target accuracy of 80\% in FL comparison.}
\label{table:cf/n label-swapped non-iid}
\begin{tabularx}{\linewidth}{ccXXXX}
\toprule
\multicolumn{1}{l}{} & \multicolumn{1}{l}{} & \multicolumn{2}{c}{test acc}                    & \multicolumn{2}{c}{\% clients}       \\
cf                   & n                    & post cluster           & 50                       & post cluster           & 50          \\
\midrule
\multirow{4}{*}{0.1}                  & 1                    & 98.2 (1.3x)          & 98.7 (1.3x)          & 49.0 (--)            & 61.0 (--) \\
                        & 3                    & 98.1 (1.3x)          & 99.0 (1.3x)          & 46.0 (--)            & 74.0 (--) \\
                        & 5                    & 98.4 (1.3x)          & 99.0 (1.3x)          & 48.0 (--)            & 73.0 (--) \\
                        & 10                   & \textbf{98.6 (1.4x)} & \textbf{99.0 (1.3x)} & \textbf{63.0 (5.3x)} & 75.0 (--) \\
\midrule
\multirow{4}{*}{0.2}                  & 1                    & 98.2 (1.3x)          & 98.6 (1.2x)          & 48.0 (--)            & 64.0 (--) \\
                        & 3                    & 98.2 (1.3x)          & 99.1 (1.3x)          & 43.0 (--)            & 77.0 (--) \\
                        & 5                    & 98.4 (1.3x)          & 99.1 (1.3x)          & 52.0 (--)            & \textbf{80.0 (--)} \\
                        & 10                   & 98.6 (1.2x)          & 99.0 (1.3x)          & 57.0 (--)            & 77.0 (--) \\
\midrule
\multirow{4}{*}{0.5}                  & 1                    & 98.1 (1.3x)          & 98.5 (1.2x)          & 49.0 (--)            & 54.0 (--) \\
                        & 3                    & 98.3 (1.3x)          & 99.0 (1.2x)          & 51.0 (--)            & 73.0 (--) \\
                        & 5                    & 98.5 (1.3x)          & 99.1 (1.2x)          & 54.0 (--)            & 76.0 (--) \\
                        & 10                   & 98.6 (1.2x)          & 99.2 (1.2x)          & 61.0 (--)            & 79.0 (--) \\
\midrule
\multirow{4}{*}{1.0}                    & 1                    & 98.3 (1.3x)          & 98.4 (1.2x)          & 53.0 (--)            & 56.0 (--) \\
                        & 3                    & 98.3 (1.3x)          & 99.0 (1.2x)          & 54.0 (--)            & 74.0 (--) \\
                        & 5                    & 98.5 (1.3x)          & 99.0 (1.2x)          & 59.0 (--)            & 76.0 (--) \\
                        & 10                   & 98.6 (1.2x)          & 99.0 (1.2x)          & 59.0 (--)            & 74.0 (--)\\
\bottomrule
\end{tabularx}
\end{table}

\begin{figure}[]
    \centering
    \includegraphics[width=\linewidth]{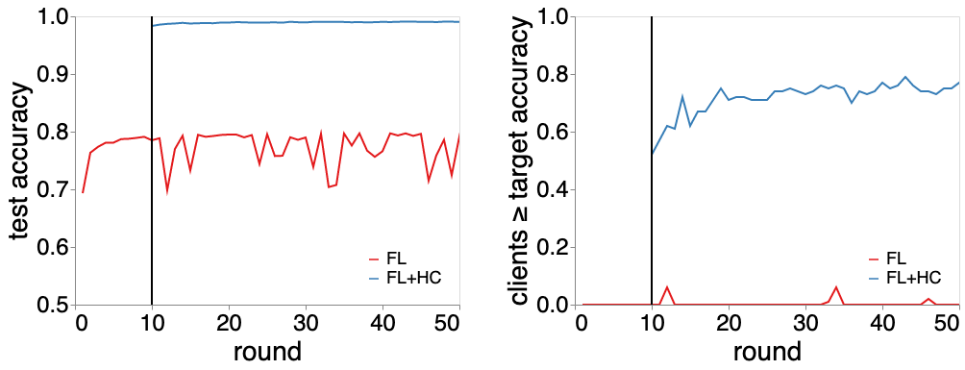}
    \caption{Label-swapped non-iid setting, using client fraction = 0.2 and number of rounds prior to clustering = 10.}
    \label{fig:cf/n r10 label-swapped non-iid}
\end{figure}

In the label-swapped non-iid setting, we know how the client datasets are clustered due to the way the data has been partitioned. We therefore expect FL+HC to find the 4 clusters during the cluster step and produce a specialised model for each. Additionally, due to the way the data is partitioned, FL can only achieve a maximum test set accuracy of 80\% in this setting due to the conflicting training objectives of each cluster. \autoref{table:cf/n label-swapped non-iid} shows that for all combinations of client fraction and number of rounds prior to clustering, FL+HC performs significantly better than FL on this data. In fact, the average test set accuracy directly after the clustering step is within 1\% of that of the iid setting and by round 50, is within 0.1\% in most cases (iid test set accuracy at round 50 is ~99.1\% for all client fractions under FL). As FL can only achieve an average test set accuracy of 80\%, in most cases under the label-swapped non-iid setting, the percentage of clients achieving the target accuracy of 99\% was 0. FL+HC was able to train up to 80\% of clients to 99\% test set accuracy. \autoref{fig:cf/n r10 label-swapped non-iid} shows how training evolves with increasing communication round for this non-iid setting. Increasing the number of rounds prior to clustering, results in improving numbers of clients reaching the target accuracy. Varying the client fraction appears to have a negligible effect on any of the metrics we recorded.

The FEMNIST non-iid setting represents a more difficult challenge in that there is an expanded label set to classify and an uneven number of samples on each client. Additionally, the number of samples available to each client is significantly fewer than in the MNIST experiments. \autoref{table:cf/n FEMNIST non-iid} details the experimental results for varying client fraction and number of rounds prior to the clustering step for this non-iid setting.

\begin{table}[]
\caption{FL+HC FEMNIST non-iid setting - cf=1, n=3 failed to complete in the time allowed.}
\label{table:cf/n FEMNIST non-iid}
\begin{tabularx}{\linewidth}{ccXXXX}
\toprule
\multicolumn{1}{l}{} & \multicolumn{1}{l}{} & \multicolumn{2}{c}{test acc}           & \multicolumn{2}{c}{\% clients}                   \\
cf                   & n                    & post cluster  &                        & post cluster            & 50                     \\
\midrule
\multirow{4}{*}{0.1} & 1                    & 30.7 (2.8x) & 74.3 (1.0x)          & 14.2 (--)             & 44.4 (0.9x)          \\
                        & 3                    & 51.4 (1.4x) & 73.6 (0.9x)          & 15.0 (27.5x)          & 42.2 (0.9x)          \\
                        & 5                    & 69.6 (1.4x) & \textbf{77.6 (1.0x)} & 29.7 (6.4x)           & \textbf{51.2 (1.1x)} \\
                        & 10                   & 71.0 (1.1x) & 76.3 (1.0x)          & 32.4 (2.0x)           & 47.1 (1.0x)          \\
\midrule
\multirow{4}{*}{0.2} & 1                    & 12.7 (0.6x) & 32.3 (0.4x)          & 0.5 (--)              & 19.6 (0.4x)          \\
                        & 3                    & 71.2 (1.9x) & 77.1 (1.0x)          & \textbf{38.7 (28.4x)} & 48.8 (1.0x)          \\
                        & 5                    & 69.4 (1.3x) & 77.4 (1.0x)          & 31.3 (4.8x)           & 49.3 (1.0x)          \\
                        & 10                   & 6.3 (0.1x)  & 5.1 (0.1x)           & 0.0 (--)              & 0.0 (--)             \\
\midrule
\multirow{4}{*}{0.5} & 1                    & 43.5 (2.2x) & 75.2 (1.0x)          & 15.0 (--)             & 46.0 (0.9x)          \\
                        & 3                    & 57.6 (1.5x) & 17.2 (0.2x)          & 28.9 (15.1x)          & 8.2 (0.2x)           \\
                        & 5                    & 69.3 (1.3x) & 77.7 (1.0x)          & 28.1 (4.7x)           & 51.5 (1.0x)          \\
                        & 10                   & 69.3 (1.0x) & 75.5 (1.0x)          & 29.7 (1.4x)           & 45.5 (0.9x)          \\
\midrule
\multirow{4}{*}{1}   & 1                    & \textbf{72.3 (2.9x)}     & 77.7 (1.0x)          & 39.2 (0.0x)           & 52.0 (1.0x)          \\
                        & 3                    & --            & --                     & --                      & --                     \\
                        & 5                    & 70.7 (1.3x) & 77.9 (1.0x)          & 31.1 (4.8x)           & 51.2 (1.0x)          \\
                        & 10                   & 69.6 (1.0x) & 75.4 (1.0x)          & 33.5 (1.4x)           & 45.0 (0.9x)         \\
\bottomrule
\end{tabularx}
\end{table}

\begin{figure}[h]
    \centering
    \includegraphics[width=\linewidth]{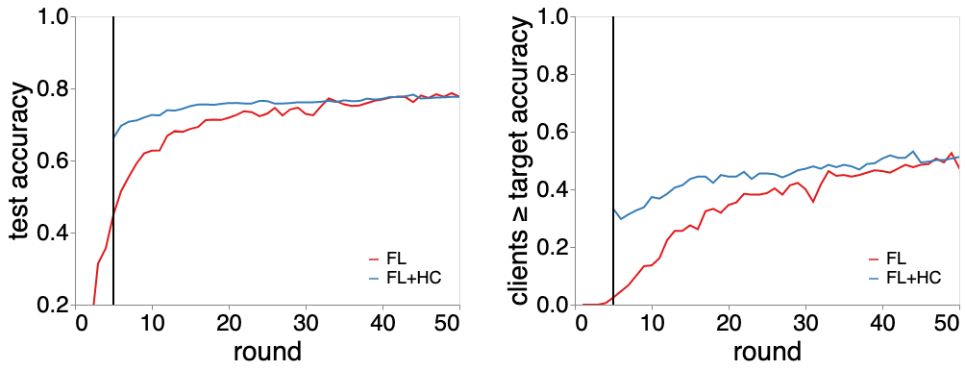}
    \caption{FEMNIST non-iid setting, using client fraction = 0.1 and number of rounds prior to clustering = 5.}
    \label{fig:cf/n r5 femnist non-iid}
\end{figure}

In most of the experiments, the clustering step has a slightly negative effect on the test accuracy and the percentage of clients reaching the target accuracy compared to FL. However in some cases there is a drastic reduction in these metrics, where the clustering step is clearly creating clusters of dissimilar clients. In general, increasing client fraction has a positive effect on the final test accuracy, indicating that FL+HC benefits from clustering on a joint model that has been trained on more data. Clustering after more rounds of FL, also seems to be beneficial in this setting.

Overall, varying the client fraction seems to have only a small (but positive) effect on test set accuracy or the percentage of clients reaching the target test set accuracy under FL+HC. However there is evidence that increasing the number of rounds prior to the clustering step can result in greater gains over FL. This is most clear in the label-swapped non-iid setting. Additionally, as shown in the figures for the pathological and label-swapped non-iid settings, FL+HC allows more clients to reach the target accuracy in any given communication round.

\subsection{Effect of varying hierarchical clustering hyperparameters}
The following experiments fix the number of communication rounds prior to clustering at 10 and the fraction of clients participating in each round of FL at 0.2. In this section, we discuss how varying the hyperparameters of the hierarchical clustering algorithm affects our metrics. We also experiment with a range of distance thresholds and report the best performing result for each combination of hyperparameters on each data partitioning scheme.

In the iid setting, the Euclidean and cosine distance metrics correctly failed to split the clients during the clustering step, resulting in a single cluster representing all the clients. As training continues, the end result is identical to federated learning. The Manhattan distance metric does cluster the clients into subgroups, resulting in reduced overall test accuracy and percentage of clients reaching the target accuracy. However, data stored across clients in the real world is unlikely to be perfectly iid.

\begin{figure}[]
    \centering
    \includegraphics[width=\linewidth]{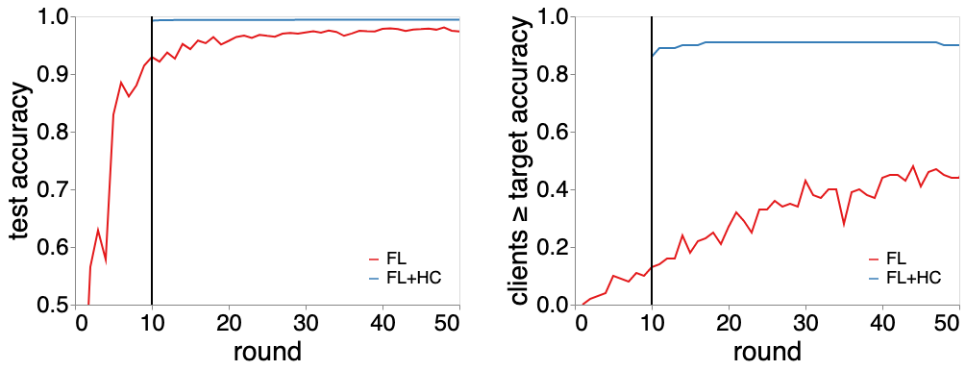}
    \caption{Pathological non-iid setting, using client fraction = 0.2 and number of rounds prior to clustering = 10, Manhattan distance metric and complete linkage.}
    \label{fig:hc r10 pathological non-iid}
\end{figure}

In the pathological non-iid setting (\autoref{table:hc pathological non-iid}), all combinations of hyperparameters yield a final test accuracy similar to, or better than training a single joint model. By round 50, the number of clients reaching the target test accuracy of 99\% is significantly better in FL+HC than FL (1.2x to 2.1x). The Manhattan distance metric performs the best overall, followed by the Euclidean distance and finally the cosine distance. Using the Manhattan distance metric, the test accuracy on this pathological non-iid setting directly after the clustering step (round 10) exceeds FL at round 50 on iid data. This represents a vast reduction ($>$5x) in communication rounds for this type of non-iid data as shown in \autoref{fig:hc r10 pathological non-iid}. Similarly, the number of clients reaching the target accuracy is greatest under the Manhattan distance metric. As the Manhattan distance metric is preferable for measuring distances between sparse high dimensional vectors \cite{Aggarwal:2001tg}, this clearly produces good clusters to train further specialised models from. The linkage method used with each distance metric appears to have a negligible effect on performance under this non-iid setting.

\begin{table}[]
\caption{FL+HC pathological non-iid setting - dist is the distance metric used judge similarity between client clusters, link is the linkage method used to determine how clients in each cluster are used in the similarity judgement.}
\label{table:hc pathological non-iid}
\begin{tabularx}{\linewidth}{llXXXX}
\toprule
\multicolumn{1}{l}{}    & \multicolumn{1}{l}{} & \multicolumn{2}{c}{test acc}  & \multicolumn{2}{c}{\% clients} \\
dist                    & link              & post cluster  & 50            & post cluster   & 50            \\
\midrule
\multirow{3}{*}{cos.} & ave.              & 89.3 (1.0x) & 97.8 (1.0x) & 12.0 (0.9x)  & 55.0 (1.3x) \\
                        & comp.             & 91.3 (1.0x) & 97.8 (1.0x) & 23.0 (1.6x)  & 56.0 (1.3x) \\
                        & sing.               & 92.0 (1.0x) & 97.7 (1.0x) & 19.0 (1.4x)  & 51.0 (1.2x) \\
\midrule
\multirow{4}{*}{L2}     & ave.              & 95.5 (1.0x) & 98.9 (1.0x) & 69.0 (4.9x)  & 86.0 (2.0x) \\
                        & comp.             & 96.6 (1.0x) & 98.8 (1.0x) & 72.0 (5.1x)  & 84.0 (1.9x) \\
                        & sing.               & 93.6 (1.0x) & 98.1 (1.0x) & 44.0 (3.1x)  & 67.0 (1.5x) \\
                        & ward                 & 97.3 (1.1x) & \textbf{99.5 (1.0x)} & 79.0 (5.6x)  & 89.0 (2.0x) \\
\midrule
\multirow{3}{*}{L1}     & ave.              & \textbf{99.4 (1.1x)} & 99.4 (1.0x) & 89.0 (6.4x)  & 91.0 (2.1x) \\
                        & comp.             & 99.2 (1.1x) & 99.3 (1.0x) & \textbf{91.0 (6.5x)}  & \textbf{92.0 (2.1x)} \\
                        & sing.               & 99.4 (1.1x) & 99.4 (1.0x) & 87.0 (6.2x)  & 91.0 (2.1x)\\
\bottomrule
\end{tabularx}
\end{table}

Under the label-swapped non-iid setting (\autoref{table:hc label-swapped non-iid}), FL+HC outperforms FL as expected when encountering clients with competing optimisation objectives. In this setting, experiments using the cosine distance outperform those using Euclidean and Manhattan distance metrics. This shows that the magnitude of the client update vectors has less of an effect than their direction in determining the correct clusters in this setting. As there are clear real clusters of clients in this non-iid setting (clients who have the same labels swapped), this poses a very different task to the pathological non-iid setting. Under all these experiments FL+HC discovers clusters of clients such that the test accuracy approaches that of FL on iid data, but only those experiments using the cosine distance metric reach the same performance. Again, in this non-iid setting, the linkage method has much less of an effect on the metrics we measure. The evolution of test accuracy and number of clients reaching the test accuracy are very similar to \autoref{fig:cf/n r10 label-swapped non-iid} under this non-iid setting.

\begin{table}[]
\caption{FL+HC label-swapped non-iid setting - dist=cosine, linkage=single failed to complete in the time allowed.}
\label{table:hc label-swapped non-iid}
\begin{tabularx}{\linewidth}{llXXXX}
\toprule
\multicolumn{1}{l}{}    & \multicolumn{1}{l}{} & \multicolumn{2}{c}{test acc}  & \multicolumn{2}{c}{\% clients} \\
dist                    & link              & post cluster  & 50            & post cluster   & 50            \\
\midrule
\multirow{3}{*}{cos.} & ave.             & \textbf{98.7 (1.3x)} & \textbf{99.1 (1.3x)} & \textbf{62.0 (--)}    & \textbf{79.0 (--)}   \\
                        & comp.             & 98.5 (1.2x) & 99.1 (1.3x) & 56.0 (--)    & 78.0 (--)   \\
                        & sing.              & --       & --       & --        & --       \\
\midrule
\multirow{4}{*}{L2}     & ave.             & 98.2 (1.2x) & 98.3 (1.2x) & 50.0 (--)    & 53.0 (--)   \\
                        & comp.           & 98.4 (1.2x) & 99.1 (1.3x) & 53.0 (--)    & 76.0 (--)   \\
                        & sing.             & 98.2 (1.2x) & 98.3 (1.2x) & 47.0 (--)    & 51.0 (--)   \\
                        & ward                 & 98.5 (1.2x) & 99.1 (1.3x) & 57.0 (--)    & 77.0 (--)   \\
\midrule
\multirow{3}{*}{L1}     & ave.             & 98.3 (1.2x) & 98.3 (1.2x) & 51.0 (--)    & 50.0 (--)   \\
                        & comp.             & 98.5 (1.2x) & 98.6 (1.2x) & 58.0 (--)    & 59.0 (--)   \\
                        & sing.              & 98.3 (1.2x) & 98.4 (1.2x) & 48.0 (--)    & 53.0 (--)  \\
\bottomrule
\end{tabularx}
\end{table}

\begin{table}[]
\caption{FL+HC FEMNIST non-iid setting}
\label{table:hc FEMNIST non-iid}
\begin{tabularx}{\linewidth}{llXXXX}
\toprule
\multicolumn{1}{l}{}  & \multicolumn{1}{l}{} & \multicolumn{2}{c}{test acc}                    & \multicolumn{2}{c}{\% clients}                  \\
dist                  & linkage              & post cluster           & 50                     & post cluster           & 50                     \\
\midrule
\multirow{3}{*}{cos.} & ave.                 & 64.2 (1.0x)          & 66.3 (0.8x)          & 19.6 (0.9x)          & 20.2 (0.4x)          \\
                        & comp.                & 62.6 (1.0x)          & 64.3 (0.8x)          & 19.3 (0.9x)          & 24.5 (0.5x)          \\
                        & sing.                & 65.4 (1.0x)          & 74.2 (0.9x)          & 19.1 (0.9x)          & 40.1 (0.8x)          \\
\midrule
\multirow{4}{*}{L2}   & ave.                 & 67.0 (1.0x)          & 78.2 (1.0x)          & 22.3 (1.0x)          & 48.2 (1.0x)          \\
                        & comp.                & 66.8 (1.0x)          & \textbf{78.7 (1.0x)} & 23.4 (1.1x)          & 52.3 (1.0x)          \\
                        & sing.                & 66.4 (1.0x)          & 77.0 (1.0x)          & 22.1 (1.0x)          & 47.7 (1.0x)          \\
                        & ward                 & \textbf{72.2 (1.1x)} & 77.4 (1.0x)          & 34.6 (1.6x)          & 48.8 (1.0x)          \\
\midrule
\multirow{3}{*}{L1}   & ave.                 & 66.5 (1.0x)          & 77.4 (1.0x)          & 22.9 (1.1x)          & 48.0 (1.0x)          \\
                        & com.                 & 69.9 (1.1x)          & 76.5 (1.0x)          & \textbf{36.0 (1.7x)} & \textbf{54.0 (1.1x)} \\
                        & sing.                & 66.9 (1.0x)          & 78.7 (1.0x)          & 23.4 (1.1x)          & 51.2 (1.0x)         \\
\bottomrule
\end{tabularx}
\end{table}

\begin{figure}[]
    \centering
    \includegraphics[width=\linewidth]{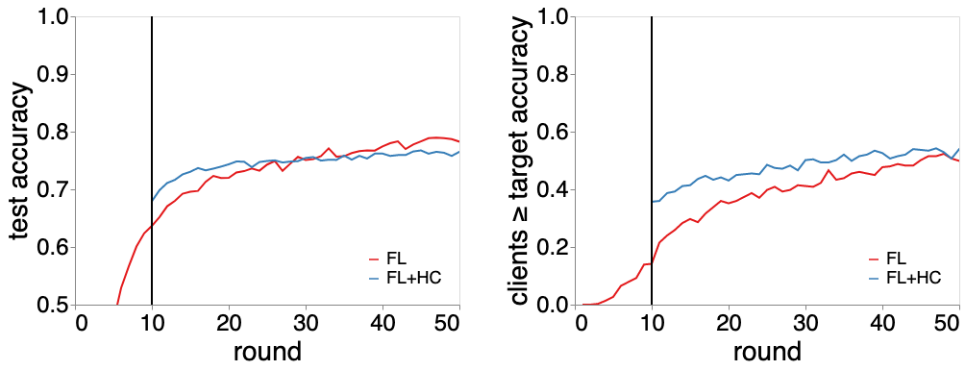}
    \caption{FEMNIST non-iid setting, using client fraction = 0.2 and number of rounds prior to clustering = 10, Manhattan distance metric and complete linkage.}
    \label{fig:hc r10 femnist non-iid}
\end{figure}

Our final batch of experiments tests the effect of varying the distance metric and the linkage mechanism in the FEMNIST non-iid setting. The results of these experiments are listed in \autoref{table:hc FEMNIST non-iid}. We see very little improvement in test accuracy or number of clients reaching the target accuracy by using FL+HC over FL. Where the cosine distance metric is employed, we see a degradation in performance. In this non-iid setting, the Euclidean distance metric produces the best final test accuracy when combined with complete linkage. The Manhattan distance metric combined with the complete distance metric (\autoref{fig:hc r10 femnist non-iid}) allows for the most clients to reach the target accuracy (1.1x more than FL by round 50). Once again, the choice of linkage mechanism has less of an effect on performance compared to the choice of distance metric.

Overall, by adjusting the hyperparameters of the hierarchical clustering algorithm, we are able to achieve greater test accuracy and an increase in the number of clients reaching a given test accuracy. Where the non-iid setting is defined by differing conditional distributions among clients (as in the label-swapped non-iid setting) the cosine distance metric performs marginally better than other distance metrics. A more natural partitioning of data (by that of the user who wrote the character in the FEMNIST non-iid setting) presents a more difficult challenge and FL+HC provides little advantage over ordinary FL. A good default setting for cases where the nature of how data is distributed among clients is unknown is to use the Manhattan distance metric, combined with the complete linkage method.

\subsection{Future work}
We have shown how adding a clustering step into the FL process can help to increase performance in some non-iid settings by inspecting the full and clean weight updates from individual clients. However, popular methods for increasing privacy (e.g. differential privacy) use noisy client updates. Future work might look at the effect of this noise on the ability of FL+HC to find good clusterings of clients. Additionally, future work might explore the effect of compression methods (designed to reduce the payload of client updates) on FL+HC. In this work, we have shown that FL+HC works well for non-iid clients using a simple CNN to classify handwritten digits. Although we are confident this method will scale to larger networks and datasets, further work is required to confirm this. Finally, an area we did not cover is FL in the presence of adversaries. A investigation of how FL+HC could used to identify malicious clients could be a promising avenue of research.
\section{Conclusion}
In this work we have presented federated learning with hierarchical clustering (FL+HC) - a novel method for training specialised machine learning models over distributed datasets. This is achieved by introducing a clustering step in the FL protocol which clusters clients according to the similarity of their model weight updates. 

We have shown results for an image classification problem using a CNN trained on simulated distributed data partitioned in a variety of iid and non-iid settings. These results show the efficacy of our approach compared to FL alone. Our first experiments varied the fraction of clients (cf) participating in each round of training and the number of rounds prior to the clustering step (n) of FL+HC. FL+HC provides results identical to FL under iid data for all but n=1. In 2 of 3 of our non-iid settings, FL+HC allows learning to converge more quickly and allows for more clients (up to 2x) to reach a target accuracy at the end of training. A second range of experiments tested the effect of varying the hyperparameters of the hierarchical clustering algorithm. Results among the non-iid settings show that FL+HC can result in a reduction in communication rounds by $>$5x when using the manhattan distance metric. Different distance metrics result in better performance depending the non-iid nature of the data. Finally we presented a recommendation for the best default settings when training on non-iid data with FL+HC.

\bibliographystyle{IEEEtran}
\bibliography{IEEEabrv,main}

\end{document}